\documentclass[letterpaper, 10 pt, conference]{ieeeconf}  
\IEEEoverridecommandlockouts                              
\usepackage{graphicx} 
\usepackage{amsmath} 
\usepackage{amssymb}  
\usepackage{multirow}
\usepackage{color}
\usepackage{subfig}
\usepackage{balance}

\title{\LARGE \bf
Memory Integrity of CNNs for Cross-Dataset Facial Expression Recognition}

\author{Dylan C. Tannugi$^{1}$, Alceu S. Britto Jr.$^{2}$ and Alessandro L. Koerich$^{1}$
\thanks{}
\thanks{$^{1}$Dylan C. Tannugi and Alessandro L. Koerich are with Department of Software and IT Engineering, \'{E}cole de Technologie Sup\'{e}rieure, University of Qu\'{e}bec, H3C 1K9, Montr\'{e}al, QC, Canada.
        {\tt\footnotesize dylancotan@gmail.com, alessandro.koerich@etsmtl.ca}}%
\thanks{$^{2}$Alceu S. Britto Jr. is with Pontifical Catholic University of Paran\'{a}, Curitiba, PR, 80215-901, Brazil.
{\tt\footnotesize alceu@ppgia.pucpr.br}}%
}

\begin{document}
\maketitle
\thispagestyle{empty}
\pagestyle{empty}

\begin{abstract}
Facial expression recognition is a major problem in the domain of artificial intelligence. One of the best ways to solve this problem is the use of convolutional neural networks (CNNs). However, a large amount of data is required to train properly these networks but most of the datasets available for facial expression recognition are relatively small. A common way to circumvent the lack of data is to use CNNs trained on large datasets of different domains and fine-tuning the layers of such networks to the target domain. However, the fine-tuning process does not preserve the memory integrity as CNNs have the tendency to forget patterns they have learned. In this paper, we evaluate different strategies of fine-tuning a CNN with the aim of assessing the memory integrity of such strategies in a cross-dataset scenario. A CNN pre-trained on a source dataset is used as the baseline and four adaptation strategies have been evaluated: fine-tuning its fully connected layers; fine-tuning its last convolutional layer and its fully connected layers; retraining the CNN on a target dataset; and the fusion of the source and target datasets and retraining the CNN. Experimental results on four datasets have shown that the fusion of the source and the target datasets provides the best trade-off between accuracy and memory integrity.
\end{abstract}


\section{INTRODUCTION}
Automatic facial expression recognition has been investigated in the last years due to the great number of applications, ranging from human-computer interaction, emotion analysis \cite{Cardinal2015} to detection of driver fatigue \cite{Zavaschi2013}. Several approaches based on handcrafted features \cite{Cossetin2016,Oliveira2011,Zavaschi2013,Zavaschi2011} such as textural, eigenfaces, etc. have been shown to be very effective. However, the most recent approaches are based on deep convolutional neural networks (CNNs) \cite{Li2018a,Lopes2015,Mousavi2016,Wang2018,Yan2018,Yang2017,Yu2015,Zavarez2017}. However, one of the main drawbacks is the amount of data required to train such deep networks. One way to circumvent the lack of data is to use CNNs pre-trained in large datasets even if the datasets and the problem lies in other domains. For instance, CNNs pre-trained in datasets such as ImageNet have been used in many different problems such as histopathologic image classification \cite{Matos2019}, environmental sound classification based on 2D representations \cite{Esmailpoor2019}, logo classification \cite{Wiggers2018} and many others. This strategy has also been used for facial expression recognition given the small size of datasets. Li and Deng \textit{et al.} \cite{Li2018} present a comprehensive survey on deep facial expression recognition (FER) that describes the standard pipeline of a deep FER system, the available datasets that are widely used in the literature. They also review existing novel deep neural networks and related training strategies that are designed for FER based on both static images and dynamic image sequences. Lopes \textit{et al.} \cite{Lopes2015} present a CNN model for facial expression classification. A large amount of grayscale face images are provided as well as the class and location of the eyes on each image. Several pre-processing steps such as spatial normalization, data augmentation by generating synthetic samples, resizing and intensity normalization are used to eliminate the bias related to the brightness of the image. The experiments carried out on the CK dataset achieved a percentage of 93.74\% of correct classification. Furthermore, the authors show that making changes to the images before providing them to a CNN improves performance. Sun \textit{et al.} \cite{Sun2016} explore audio, visual and physiological signals to continuously predict the value of the emotion dimensions (arousal and valence). They have evaluated a variety of handcrafted and deep visual features such as Dense SIFT features (MSDF), and some types of Convolutional neural networks (CNNs) features to recognize the expression phases of the frames. Yu and Zhang \textit{et al.} \cite{Yu2015} present an approach for facial expression recognition for the Emotion Recognition in the Wild Challenge 2015. The proposed approach is based on an ensemble of three state-of-the-art face detectors, followed by a classification module with the ensemble of multiple deep convolutional neural networks (CNN). Each CNN model is initialized randomly and pre-trained on FER dataset. The pre-trained models are then fine-tuned on the training set of SFEW 2.0 and achieves 55.96\% and 61.29\% on the validation and test set of SFEW 2.0, respectively, surpassing the baseline in 35.96\% and 39.13\%, respectively.

Zhang \textit{et al.} \cite{Zhang2019} present a comprehensive review of transfer-learning methods for cross-dataset visual recognition. In particular, we are interested in homogeneous feature and label spaces, where the feature spaces and label spaces of the source and the target datasets are identical, but domain divergence exists across the source and the target datasets. In this problem, a small number of labeled data in the target domain is available. However, the labeled target data are generally insufficient for learning an effective classifier. This is also called supervised domain adaptation or few-shot domain adaptation in the literature \cite{Zhang2019}. Another group of methods transfers the parameters of discriminative classifiers across datasets \cite{Xu2014,Yang2007,Jiang2008}. Mousavi \textit{et al.} \cite{Mousavi2016} use deconvolution to visualize the characteristics learned by a CNN trained with the Cohn-Kanade (CK) dataset and tested on the same dataset and on the JAFFE dataset. Zavarez \textit{et al.} \cite{Zavarez2017} propose a CNN based on the VGG net \cite{Simonyan2014} which uses two different initialization methods: randomly initialized weights and weights from a pre-trained VGG-Face model. Experiments were carried out with different combinations of training and test sets of seven datasets: AR Face, CK+, BU3DFE, JAFFE, MMI, RaFD, and KDEF. For each combination, the training set is composed of the merging of six datasets leaving one dataset for the test set. The fine-tuned VGG net performed better than the VGG initialized with random weights and achieved the accuracy of 88.58\% on CK+ as the test set. Indeed, only the combination having JAFFE as test set obtained a worse result for the fine-tuned VGG net. This shows that fine-tuning a CNN is often effective as well as more stable than one initialized with random weight. The fine-tuned model also outperformed the state-of-the-art for four out of five datasets evaluated. Mayer \textit{et al.} \cite{Mayer2014} present a system for facial expression recognition that is evaluated on multiple datasets. Comparing classifiers across datasets determines the classifiers' capability to generalize more reliable than traditional self-classification. Li and Deng \textit{et al.} \cite{Li2018a} present a deep Emo-transfer network (DETN) to deal with the problem of cross-dataset facial expression recognition which embeds maximum mean discrepancy in the deep architecture to reduce dataset bias. The experimental results on both lab-controlled and real-world facial expression datasets have shown competitive performances across various facial expression transfer tasks. Wang \textit{et al.} \cite{Wang2018} introduce an unsupervised domain adaptation method, which is especially suitable for small unlabeled target datasets. They have trained a generative adversarial network (GAN) on the target dataset and use the GAN generated samples to fine-tune the model pre-trained on the source dataset. In the process of fine-tuning, the unlabeled GAN generated samples distributed pseudolabels dynamically according to the current prediction probabilities. They demonstrated the effectiveness of the method on four facial expression recognition datasets with two CNN structures. Yang \textit{et al.} \cite{Yang2017} present a method based on transfer learning and sparse coding to learn a dictionary and transfer it to facial expression to obtain a feature representation by sparse coding. The experimental results in CK+, JAFFE, and NVIE show that the transfer learning based on sparse coding method can effectively improve the expression recognition rate in the cross-domain facial expression recognition task. Yan \textit{et al.}\cite{Yan2018} proposed a transductive deep transfer learning architecture based on VGGface16-Net to jointly learn a common optimal nonlinear discriminative features from the non-frontal facial expression samples between the source and target datasets and cross-dataset non-frontal facial expression classification task. Extensive cross-dataset experiments on BU-3DEF and Multi-PIE datasets are presented, and the experimental results show that the transductive deep transfer network outperforms the state-of-the-art cross-database facial expression recognition methods.

Most of the works focus on adapting an existing CNN to a new dataset (target), maximizing the performance of such a model on the target dataset and leaving aside (forgetting) the performance on the source dataset. In this paper, we evaluate different adaptation approaches with the aim of obtaining a model that provides the best trade-off in terms of performance for both the source and target datasets. In case of new subjects are added to the data, which is the best way to integrate such new users? Should we modify or not the learned model? And if we decide to modify it, should we fully train the model and on which data? Should we rebuild it from scratch or preserve the already learned parameters and just slightly adapt it to the new data? Therefore, in such a case, the main question that we are trying to answer in this paper is whether transfer learning is useful or not. How useful is transfer learning when we have several datasets for the same problem, and therefore we can combine these datasets and design a customized architecture for tackling the problem. Furthermore, are the current transfer learning strategies able to deal with memory integrity and plasticity as CNNs have the tendency to forget patterns they have learned?

\begin{figure*}[htpb!]
\centering
\includegraphics[scale=0.32]{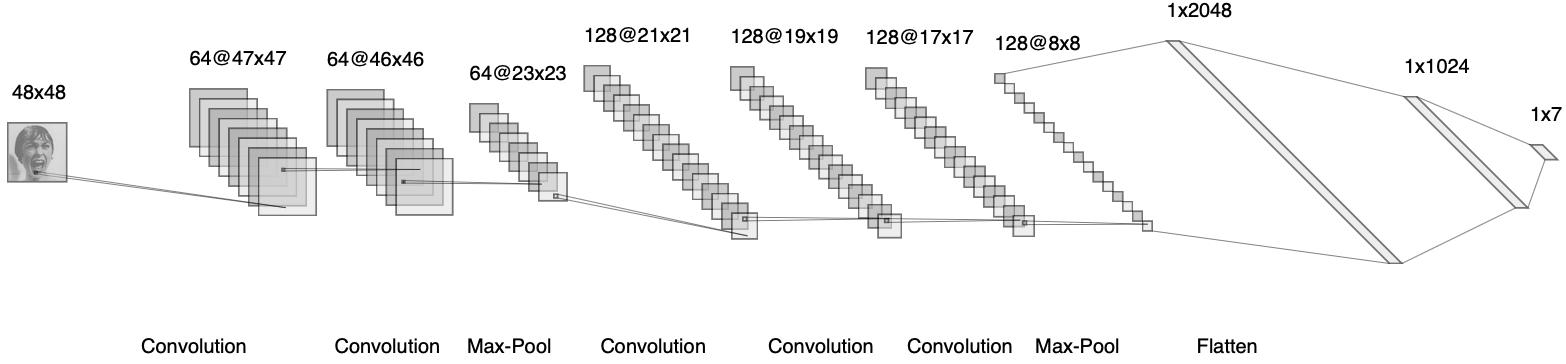}
\caption{Architecture of the baseline CNN for facial expression recognition.}
\label{fig:cnn}
\end{figure*}

In this paper, we show that transfer learning or fine-tuning of CNNs is not so effective to preserve memory integrity. Instead, fully training the network is a much better option as it usually leads to the best performance on both the source and the target datasets. The contributions of this paper are twofold: (i) characterize different transfer learning strategies subject to the constraint of preserving memory integrity; (ii) fine-tuning or retraining a CNN? Which action leads to the best performance if we look for a good balance between the performance on the source dataset and on a target dataset.

This paper is organized as follows: Section~\ref{sec:II} describes some transfer-learning strategies for cross-dataset recognition. Experimental results on cross-dataset transfer learning and recognition on four datasets are presented in Section~\ref{sec:exp}. Finally, the conclusions and perspectives of future work are stated in the last section.  




\section{Transfer Learning Strategies for Cross-Dataset Recognition}
\label{sec:II}
In facial expression recognition, most of the datasets have a small number of labeled data. However, such datasets are generally insufficient for learning an effective model. One way to circumvent such a lack of data is to use cross-dataset strategies where there are at least two datasets, one of them referred to as the source dataset is used to train a model and the other, referred to as the target dataset is the dataset to which the model has to be adapted. Deep networks can generally learn more transferable features \cite{pmlr-v27-bengio12a,pmlr-v32-donahue14}.

Considering that the source and the target datasets belong to the same domain, namely facial expression recognition, according to Zhang \textit{et al.}\cite{Zhang2019}, this can be characterized as homogeneous feature spaces and label spaces because observation spaces $\mathcal{X}$ as well as the label spaces $\mathcal{Y}$ are both identical between source and target datasets. Hence the target and the source datasets are different in terms of data distributions ($P(\mathcal{X},\mathcal{Y})$).

Therefore, in our approach for cross-dataset facial expression recognition, we consider a pre-trained CNN on a dataset that belongs to the same domain and for the same task. The architecture of the proposed CNN is shown in Fig.~\ref{fig:cnn}. This architecture is a simplified version of a VGG network \cite{Simonyan2014}, with less convolutional layers, given the amount of data available for training. However, the number of parameters is 19.2M which is about seven times less than a VGG network, which has 138M of trainable parameters. The proposed CNN takes a 48$\times$48 gray-level image as input and it has five convolutional layers interchanged with two max-pooling layers and three fully connected layers. Rectified linear units (ReLU) and batch normalization are used after each convolutional layer. The first-two convolutional layers have 64 filters of size 2$\times$2 and the last three convolutional layers have 128 filters of size 3$\times$3. Furthermore, the spatial resolution of the second and fifth convolution layers are reduced by twice with two max-pooling layers to limit the number of parameters and the computation cost. The first fully connected layer is made up of 2,048 units, followed by a second layer of 1,024 and the output layer with seven units. Drop-out of 50\% is used after the two-first fully connected layers to reduce the over-fitting. We use such a CNN in our study to compare four different approaches of crossing different datasets. For each of these approaches, only two datasets are crossed at each time, but it is also possible to cross more than two datasets.

\begin{figure}[htpb!]
\centering
{
\setlength{\tabcolsep}{0em} 
\begin{tabular}{c}
    \includegraphics[scale=0.6]{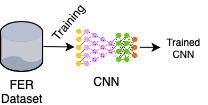}\\
    (a) \\
    \includegraphics[scale=0.6]{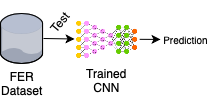}\\
 (b)  \\
\end{tabular}
}
\caption{The baseline CNN is: (a) trained on the training split of the source dataset (FER) and; (b) evaluated at test partition of the same  dataset (source dataset).}
\label{fig:fer_baseline}
\end{figure}

\subsection{Fine-tuning the Fully Connected (FC) Layers}
\begin{figure}[htpb!]
\centering
{
\setlength{\tabcolsep}{0.0em} 
\begin{tabular}{c}
    \includegraphics[scale=0.6]{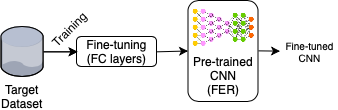}\\
    (a)  \\
      \includegraphics[scale=0.6]{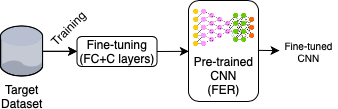}\\
    (b)  \\
      \includegraphics[scale=0.6]{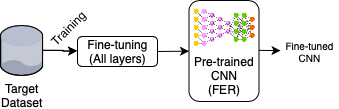}\\
    (c)
\end{tabular}
}
\caption{The baseline CNN is fine-tuned using three strategies: (a) fully connected layers (FC); (b) fully connected layers and the last convolutional layer (FC+C); (c) all layers.}
\label{fig:fr_strat}
\end{figure}

For this first approach, we consider a pre-trained CNN on the dataset that belongs to the same domain and for the same task. The CNN is therefore trained and optimized on a source dataset (FER dataset) as illustrated in Fig.~\ref{fig:fer_baseline}. A second dataset named target dataset is then used to fine-tune only the fully connected (FC) layers of the pre-trained CNN as illustrated in Fig.~\ref{fig:fr_strat}a. This new training considers the previous weights as the initial ones and it looks to minimize the error of the network on the target dataset. The parameters of the other layers are not modified. In this case, our assumption is that the representation learned in the convolutional layers on the source dataset is reasonable and general and it does not need to change because the source and target tasks are in the same domain. Usually, the high-level features computed by the last few layers are usually task-specific and are not transferable to new target tasks.
 
\subsection{Fine-tuning the Fully Connected (FC) and the Last Convolutional Layer}
This second approach also requires the use of an already trained CNN. As with the previous method, CNN was trained by the FER database. This method is also a method of adjusting some CNN layers by a second dataset as illustrated in Fig.~\ref{fig:fr_strat}b. This time, the adjusted layers are the last convolutional layer and the three fully connected layers. This second method is very similar to the first one. The difference is that by adjusting the last CL layer, we are also modifying the learned representation. The convolutional layers are responsible for extracting features that best characterize the classes. Modifying the convolutional layers, therefore, modifies the features that the model considered important. In this case, the assumption is that the representation learned in the convolutional layers is reasonable, but it can be adapted to the target dataset. 

\subsection{Fine-tuning All Layers}
The third approach also requires a pre-trained CNN. As for the two previous cases, the CNN was pre-trained on the FER database. For this approach, the CNN is fully trained a second time on a target dataset as illustrated in Fig.~\ref{fig:fr_strat}c. However, the CNN is initialized with the parameter values learned previously on the source dataset. Therefore, this approach involves training the CNN twice, with two different datasets while keeping the parameters learned on the source dataset and the initial parameters for learning on the target dataset. In this case, the assumption is that the whole network should be adapted to the target task and dataset, but it still will keep some "memory" of the source dataset.   

\subsection{Train the CNN on Merged Datasets}
The last approach does not require a pre-trained CNN. Indeed, it consists of merging the source and target datasets into a single dataset and use such a dataset to train a CNN initialized with random weights. For this approach, the FER dataset was always used as the source dataset and it was merged with one of the other three target datasets. The CNN is then trained in both datasets simultaneously. This is the most time-consuming approach as it requires full training of the network.

\begin{figure}[htpb!]
\centering
{
\setlength{\tabcolsep}{0em} 
\begin{tabular}{c}
    \includegraphics[scale=0.6]{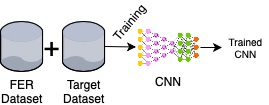}\\
\end{tabular}
}
\caption{The baseline CNN is randomly initialized and trained on merged source and target datasets.}
\label{fig:fr_strat2}
\end{figure}

\begin{figure}[htpb!]
\centering
{
\setlength{\tabcolsep}{0.0em} 
\begin{tabular}{c}
    \includegraphics[scale=0.6]{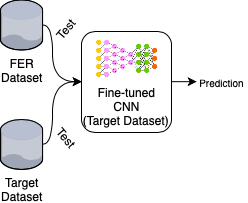}\\
\end{tabular}
}
\caption{The fine-tuned CNNs are evaluated on both source and target test datasets.}
\label{fig:fr_strat3}
\end{figure}

Given the four proposed approaches for crossing datasets, which of them is the most appropriate in a situation where our aim is to minimize the error rate on both source and target datasets? 

\section{Experimental Results}
\label{sec:exp}
We have evaluated the four proposed approaches of crossing different datasets on four publicly available datasets: FER, JAFFE, TFEID and MUG. The Facial Expression Recognition (FER) Challenge dataset is made up of 35,888 48$\times$48 gray-level images split into 28,709 images for training (80\%), 3,589 images for validation (10\%) and 3,589 images for testing (10\%). We have tried to keep the same proportion of data into each subset for the three other datasets. Besides, the data splitting has also considered that a subject should be only in one of the subsets to avoid biasing the CNNs. The Japanese Female Facial Expression (JAFFE) dataset is made up of only 213 gray-level images with 256$\times$256 pixels from 10 Japanese females. This dataset was split into 131 images for training (60\%), 15 images for validation (10\%) and 67 (30\%) for testing. The Taiwanese Facial Expression Image dataset (TFEID) is made up of 510 color images of 40 Taiwanese subjects with different dimensions, ranging from 600 to 606 of height and from 480 to 595 in width. This dataset was split into 405 images for training (79.4\%), 47 images for validation (9.2\%) and 58 (11.4\%) for testing. Finally, the Multimedia Understanding Group (MUG) Facial Expression dataset is made up of 919 color videos with between 60 and 100 frames of 896$\times$896 of resolution from 86 Caucasian adults. The videos contain a sequence of temporal phases, starting and finishing with neutral expressions with the apex of the labeled emotion in the middle of the sequence. We have selected the first and last frames as neutral expressions and the six frames in the middle of the sequence as the corresponding emotion expressed in the video. Therefore, we ended up with 7,352 images which were split into 5,877 (80\%) for training, 734 (10\%) for validation and 741 (10\%) for testing. Table \ref{tab:data} summarizes the datasets and the amount of data in each partition.

\begin{table}[htpb!]
\renewcommand{\arraystretch}{1.2}
\small
\begin{center}
\begin{tabular}{|l|r|r|r|r|}
\hline
 & \multicolumn{4}{c|}{Number of Samples and (\%)}\\
\cline{2-5}
Dataset & \multicolumn{1}{c|}{Total} & \multicolumn{1}{c|}{Training} & \multicolumn{1}{c|}{Validation} & \multicolumn{1}{c|}{Test} \\
\hline
FER & 35,888 & 28,709 (80.0) & 3,589 (10) & 3,589 (10.0) \\
JAFFE & 213 & 131 (61.5) & 15 (7.0) & 67 (31.5) \\
TFEID & 510 & 405 (79.4) & 47 (9.2) & 58 (11.4) \\ 
MUG & 7,352 & 5,877 (80.0) & 741 (10) & 734 (10.0) \\ 
\hline
\end{tabular}
\end{center}
\caption{The four datasets and the three subsets.}
\label{tab:data}
\end{table}

\begin{figure*}[htpb!]
\centering
\includegraphics[scale=0.815]{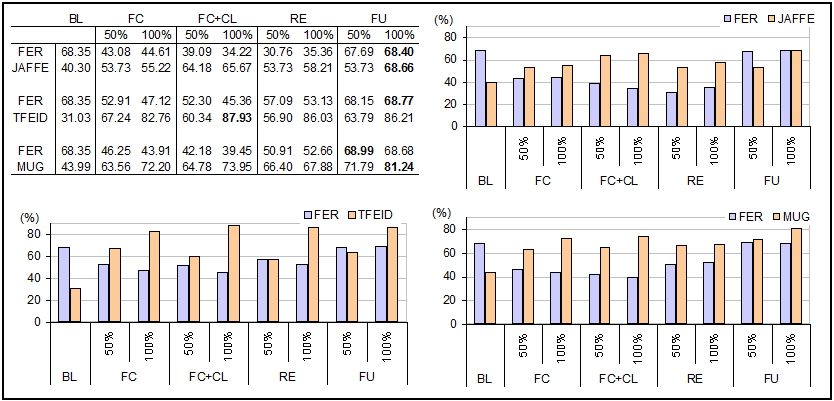}
\caption{Cross-dataset results using FER as the source dataset and the three target datasets: JAFFE, TFEID, and MUG. Learning process based on half (50\%) and the whole (100\%) training set. Accuracies of the Baseline model (BL), fine-tuning the fully connected layers (FC), fine-tuning the fully connected plus the last convolutional layer (FC+CL), retraining of all model layers (RE), and fusion of training sets (FU).}
\label{fig:Results}
\end{figure*}

Figure \ref{fig:Results} shows the results in terms of accuracy on the test set of JAFFE, TFEID, and MUG. It is possible to see that the error percentage is lower for the experiments made with the whole of the dataset than learning with half of this dataset. This suggests that the larger the dataset used for learning, the more precise the CNN classification will be for the data in this dataset. However, by observing the results for the FER test set, it is not possible to see a trend. It is therefore conceivable that the amount of data in the target training dataset does not really affect the effectiveness of the CNN classification for the source dataset.

\subsubsection{Performance on the source test set after fine-tuning the CNN:} The three fine-tuning approaches have significantly reduced the accuracy of the resulting CNN on FER test set between 20.94\% and 28.67\% in average. On the other hand, fusing FER dataset and a target dataset has kept or even improved the accuracy in 0.27\% on average.     

\subsubsection{Performance on the target test sets after fine-tuning the CNN:} The three fine-tuning approaches have significantly improved the accuracy of the resulting CNN on the target test sets between 20.57\% and 37.41\% on average. Fusing FER dataset and a target dataset has kept or even improved the accuracy in 40.26\% on average.




Figures \ref{fig:Results}(a), \ref{fig:Results}(b) and \ref{fig:Results}(c) also show that only the merging approach does not have a higher error percentage than the baseline model (BL). In some cases, such an approach even achieves slightly lower error rates. Therefore, it seems that the merging approach (FU) is the most effective regarding the performance of the resulting CNN on the source dataset. Therefore, we can conclude that the dataset fusion (FU) approach provides the best trade-off because it is the only approach that does not have a negative effect on the source dataset while it is the approach that gives the best performances for the target datasets.


\section*{Conclusion}
In this paper we have investigated several approaches to supervised domain adaptation of CNNs, considering the problem of facial expression recognition. The main drawback of fine-tuning pre-trained CNNs is that it implicitly imposes a forgetting mechanism which degrades their performance on the source dataset. Therefore, we may say that in general, fine-tuning pre-trained CNNs does not preserve the memory integrity of the model. The availability of pre-trained networks has attracted several researchers to simply fine-tune these pre-trained networks on their target datasets. However, one has to take into account that "transfer learning" is not always useful, especially when we are working on the same domain. Therefore, one of the main contributions of this paper is to show that the massive use of pre-trained networks does not always lead to the best performance. It is preferable to use a simpler architecture, more adapted to the problem than using a complex pre-trained network and transfer learning. 

Besides that, our experimental results highlight the importance of the context in training or fine-tuning CNNs. Notice that the JAFFE, TFEID and MUG datasets have 219$\times$, 71$\times$, and 5$\times$ less training samples than the source dataset, respectively. However, when the CNN is trained with the fusion of FER and JAFFE training datasets, even if the amount of JAFFE samples represents only 0.45\% of the training samples, the performance on the JAFFE test set boosted from 40.30\% to 68.66\%, an improvement of 28.36\% in accuracy. Furthermore, fusion approach is the best concerning the memory integrity, as it preserves almost the same accuracy on the source test set. On the other hand, if we fine-tune only the fully connected layer with few images, we are able to achieve an improvement of at least 13.43\% on the JAFFE test set. The same behavior was observed for the two other target datasets. However, as mentioned before, the fine-tuning has a catastrophic impact on memory integrity.

Since our goal is quite different from other researchers which aims to create the most effective facial expression recognition approach for a target domain, comparing the results achieved in this work with the results of some of the works reviewed earlier, does not seem relevant. Among the five datasets used in \cite{Zavarez2017}, only JAFFE was also used in our experiments. For such a dataset, Zavarez \textit{et al.} \cite{Zavarez2017} achieved 44.32\% of accuracy with their fine-tuned model while our approach achieved performances between 40.30\% and 68.66\% with no fine-tuning and with the fusion of training set and fully training of the CNN respectively. Therefore, we may say that the proposed approaches for dataset crossing are more efficient. Finally, Yang \textit{et al.}\cite{Yang2017} use JAFFE as the target dataset and the recognition rates achieved in cross-domain datasets are always lower than 40\%. 
 

 
 
\balance




\end{document}